\documentclass{article}
\usepackage[utf8]{inputenc}
\usepackage{ragged2e}

\usepackage{graphicx}
\graphicspath{ {images/} } 
\usepackage{array}
\newcolumntype{P}[1]{>{\raggedright}p{#1 cm}}

\usepackage[ruled,vlined]{algorithm2e}
\usepackage{array}

\usepackage{listings}
\usepackage{color}
\usepackage[hmarginratio=1:1, vmargin=1in, hmargin=1.5in, columnsep=10pt]{geometry} 
\usepackage[small,labelfont=bf,up]{caption} 
\usepackage{booktabs} 
\usepackage{float} 

\usepackage{abstract} 

\usepackage[natbib=true, style=nature, sorting=none]{biblatex}
\usepackage{amsmath}
\usepackage{amssymb}
\usepackage{titlesec} 
\titleformat{\section}{\Large\scshape\centering\bfseries}{\thesection}{1em}{}

\usepackage{graphicx,wrapfig}
\usepackage{bbm}
\usepackage{gensymb}
\usepackage{siunitx}
\usepackage{dsfont}

\usepackage[pagewise]{lineno}

\title{\textbf{Deep Distilling: automated code generation using explainable deep learning}\vspace{-2em}}

\date{}

\begin{document}

\maketitle 
\begin{center}
    Paul J. Blazek\footnotemark[1]\footnotemark[2]\footnotemark[3], Kesavan Venkatesh\footnotemark[1]\footnotemark[4], Milo M. Lin\footnotemark[1]\footnotemark[2]\footnotemark[3]\footnotemark[5]\renewcommand{\thefootnote}{\fnsymbol{footnote}}\footnote[1]{Correspondence to: milo.lin@utsouthwestern.edu}
\end{center}
\footnotetext[1]{Green Center for Systems Biology, University of Texas Southwestern Medical Center, Dallas, TX 75390}
\footnotetext[2]{Department of Bioinformatics, University of Texas Southwestern Medical Center, Dallas, TX 75390}
\footnotetext[3]{Department of Biophysics, University of Texas Southwestern Medical Center, Dallas, TX 75390}
\footnotetext[4]{Department of Biomedical Engineering, Johns Hopkins University, Baltimore, MD 21218}
\footnotetext[5]{Center for Alzheimer’s and Neurodegenerative Diseases, University of Texas Southwestern Medical \hfill \break \indent \indent \indent Center, Dallas, TX 75390}

\begin{abstract}
\noindent
Human reasoning can distill principles from observed patterns and generalize them to explain and solve novel problems. The most powerful artificial intelligence systems lack explainability and symbolic reasoning ability, and have therefore not achieved supremacy in domains requiring human understanding, such as science or common sense reasoning. Here we introduce deep distilling, a machine learning method that learns patterns from data using explainable deep learning and then condenses it into concise, executable computer code. The code, which can contain loops, nested logical statements, and useful intermediate variables, is equivalent to the neural network but is generally orders of magnitude more compact and human-comprehensible. On a diverse set of problems involving arithmetic, computer vision, and optimization, we show that deep distilling generates concise code that generalizes out-of-distribution to solve problems orders-of-magnitude larger and more complex than the training data. For problems with a known ground-truth rule set, deep distilling discovers the rule set exactly with scalable guarantees. For problems that are ambiguous or computationally intractable, the distilled rules are similar to existing human-derived algorithms and perform at par or better. Our approach demonstrates that unassisted machine intelligence can build generalizable and intuitive rules explaining patterns in large datasets that would otherwise overwhelm human reasoning.
 \vspace*{3 em}
\end{abstract}

 
Human reasoning is able to successfully discover principles from the observed world and systematize them into rule-based knowledge frameworks such as scientific theories, mathematical equations, medical treatment protocols, heuristics, chemical synthesis pathways, and computer algorithms. The automation of this reasoning process is the long-term goal of artificial intelligence (AI) and machine learning. However, there is a widely acknowledged trade-off between models that can be explained intuitively (e.g. decision trees) and models that have wide scope and high predictive accuracy (e.g. neural networks) \cite{arrieta2019explainable, darpa}. Explainability is important for legal and ethical reasons \cite{legal}, for making AI systems more amenable to modification and rational design, and for providing the guarantees and predictability needed for high-stakes applications such as medical diagnosis or autonomous motor vehicle driving. More fundamentally, human comprehensibility of automated reasoning is necessary to transition from computer-aided to computer-directed scientific discovery.
\par
A rigorous test of the explainability of any machine learning model is whether the learned model can be written as succinct computer code. This provides an unambiguous explanation of the learned algorithm, and it serves as a platform for testing performance. This is the goal of inductive programming (IP) or programming-by-example, through which computer code that infers the relationship between training data inputs and outputs is generated. To date, IP has been restricted to writing code that automates straightforward repetitive tasks involving simple inputs, easy manipulative operations, and only a few lines of code \cite{ip_realworld, approaches_ip, survey_ip, deepcoder, flashmeta}. This is due to the limitations imposed by the large space of functions through which IP methods search when generating code. IP automates the interpretation of data and learning of algorithms \textit{for} the user, which is fundamentally different from other automatic programming approaches that simply automate the translation or completion of commands \textit{from} the user \cite{program_synthesis, code_generation}. Examples of the latter include automatic code completion and debugging (e.g. Tabnine, Kite, OpenAI Copilot, etc.), as well as translating human-given sketches or textual descriptions of algorithms or applications \cite{sketching}. Notably, generating code from human-written function names and docstrings has recently been demonstrated by Copilot using the natural language processing (NLP) model GPT-3 \cite{gpt3, gpt3_code}. However, NLP-based methods learn about how a subject is described rather than about the subject itself, and are therefore not able to discover new knowledge. This also means that they are restricted by the imprecision inherent in the natural language description of nontrivial algorithms.
\par
Recently, we introduced a class of explainable deep neural networks called essence neural networks (ENNs) that are trained using a new symbolic approach \cite{enns} instead of the prevailing method of error minimization with gradient descent and back-propagation. ENNs perform well on both computer vision and logical reasoning tasks. For symbolic tasks, we previously reported that ENNs are fully interpretable and can be manually translated to pseudocode.
\par
Here we introduce deep distilling, a general inductive programming method that automatically distills training data into computer code to explain the data, first training symbolic ENNs and then translating the learned parameters to a human-understandable algorithm. Deep distilling is able to discover the underlying rules that govern well-defined systems such as cellular automata and discover generalizable algorithms for non-trivial problems from arithmetic, computer vision, and NP-hard logic problems. These results suggest a viable framework for automatic discovery of principles and algorithms from labeled data.

\section*{Results}
\subsection*{Deep distilling condenses ENNs to code}

\begin{figure}[h!]
  \begin{minipage}[t]{0.6\textwidth}
    \vspace{0.01pt}
    \includegraphics[width=\textwidth]{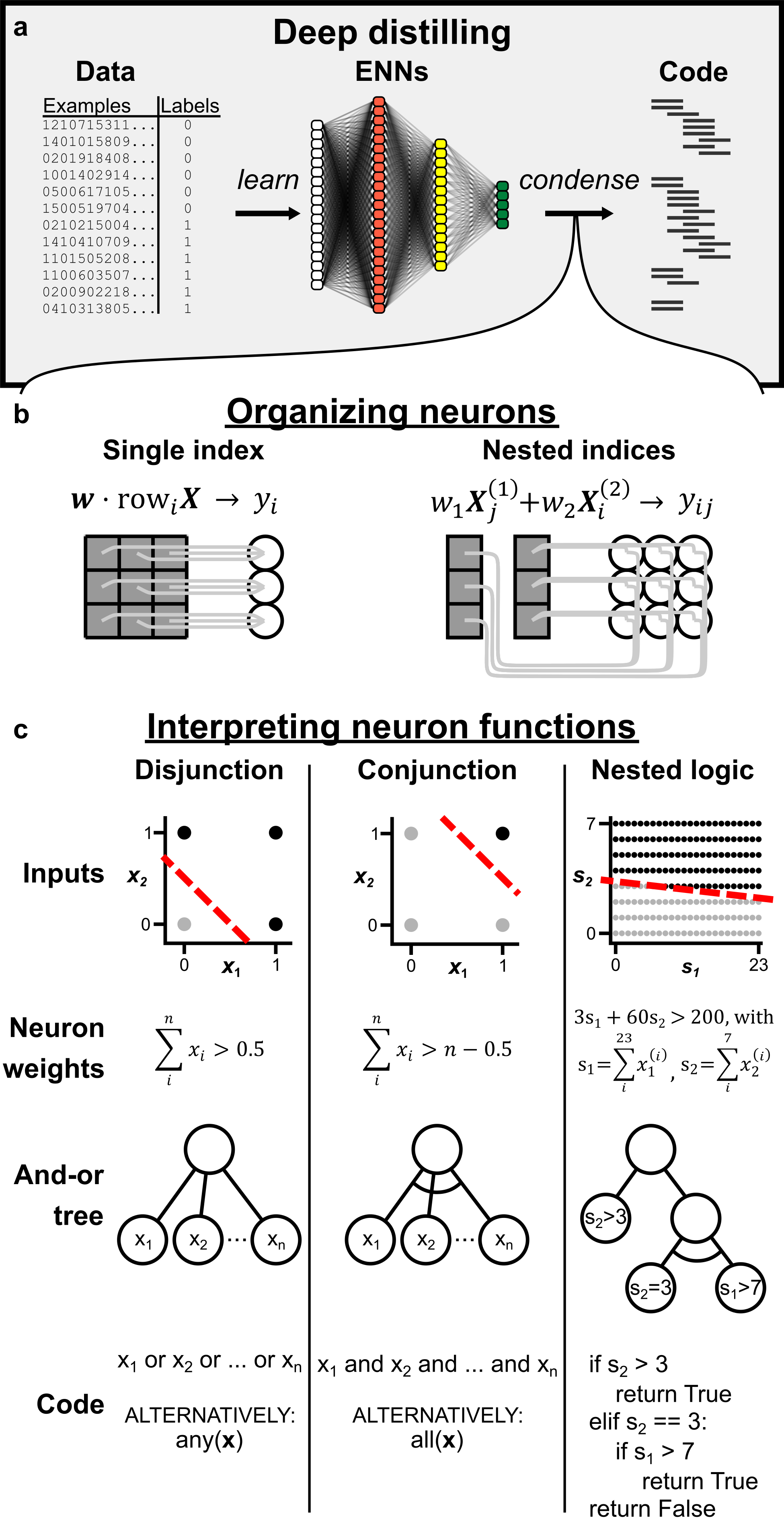}
  \end{minipage}\hfill
  \begin{minipage}[t]{0.35\textwidth}
    \vspace{0.01pt}
    \caption{
       \textbf{Deep distilling automatically writes computer code.}\textbf{(a)} Deep distilling learns from data to produce an ENN which is then condensed to computer code. ENNs are inspired by a neurocognitive model in which neurons can make relative distinctions (differentia neurons, red) or absolute distinctions (subconcept and concept neurons, yellow and green respectively). \textbf{(b)} The first step taken by the ENN condenser is to organize neurons into groups with similar connectivity patterns so they can be iterated over by a for-loop. Two examples are shown, including one which uses a nested for-loop. \textbf{(c)} The condenser's second step interprets neurons as implementing a logical function, examples of which are shown here. In each, the neuron's weights define a hyperplane (red dashed line) that distinguishes samples of discrete inputs $x_i$ (or distilled variables $s_i$). These distinctions can be understood as and-or trees or as computer code.
    }
  \end{minipage}
\end{figure}

ENNs are inherently explainable because each neuron is designed to make a specialized distinction with respect to one or two \textit{concepts}. Each neuron is either a \textit{concept neuron} that distinguishes a particular subset of training samples (i.e. a \textit{concept}) from every other sample, or a \textit{differentia neuron} that distinguishes between two particular concepts \cite{enns}. When the activation function is the sign function, this distinction is discrete (0 signifying a tie). Discrete distinctions allow each neuron output to serve as a logical symbol (i.e. +1 signifies TRUE, -1 signifies FALSE). In this way ENNs enable an automated approach to human-like reasoning.
\par
Deep distilling is a two-step process that (i) uses ENNs to learn how input data map to output values, then (ii) condenses this ENN to succinct, human-understandable code (Fig. 1a). Condensing an ENN can be divided into two steps (Methods). The first step organizes neurons with similar connectivity patterns into functional groups (Fig. 1b), in which each neuron has the same set of weights applied to different input synapses. This allows the neurons to be indexed and iterated over (Fig. 1b). The ENN condenser's second step interprets each group's set of weights as a function (Fig. 1c), as in the very first artificial neurons designed by McCulloch and Pitts \cite{mcculloch_pitts}. Two well-known examples are logical disjunction and conjunction (Fig. 1c). The condenser can also create a single \textit{distilled variable} to represent the sum of multiple inputs, and when a neuron uses multiple distilled variables, it is possible to represent the neuron's function as a set of nested logical functions (e.g. Fig. 1c). An ENN condenser can write the distilled algorithm in any desirable programming language (or even in pseudocode or prose); we have chosen here to report code written in Python.
\par
When training and condensing an ENN, the training samples all have the same input size, producing code with hard-coded constants (e.g. a for-loop's iteration range). However, when deep distilling is performed several times with training samples of different sizes, the same code can be distilled with different hard-coded constants. A function is fit to these constants, which is then substituted into the code (Methods). The generalized code is able to handle inputs of any size, even orders of magnitude larger than that of the training samples.

\begin{figure}[h!]
  \begin{minipage}[t]{0.7\textwidth}
    \vspace{0.01pt}
    \includegraphics[width=\textwidth]{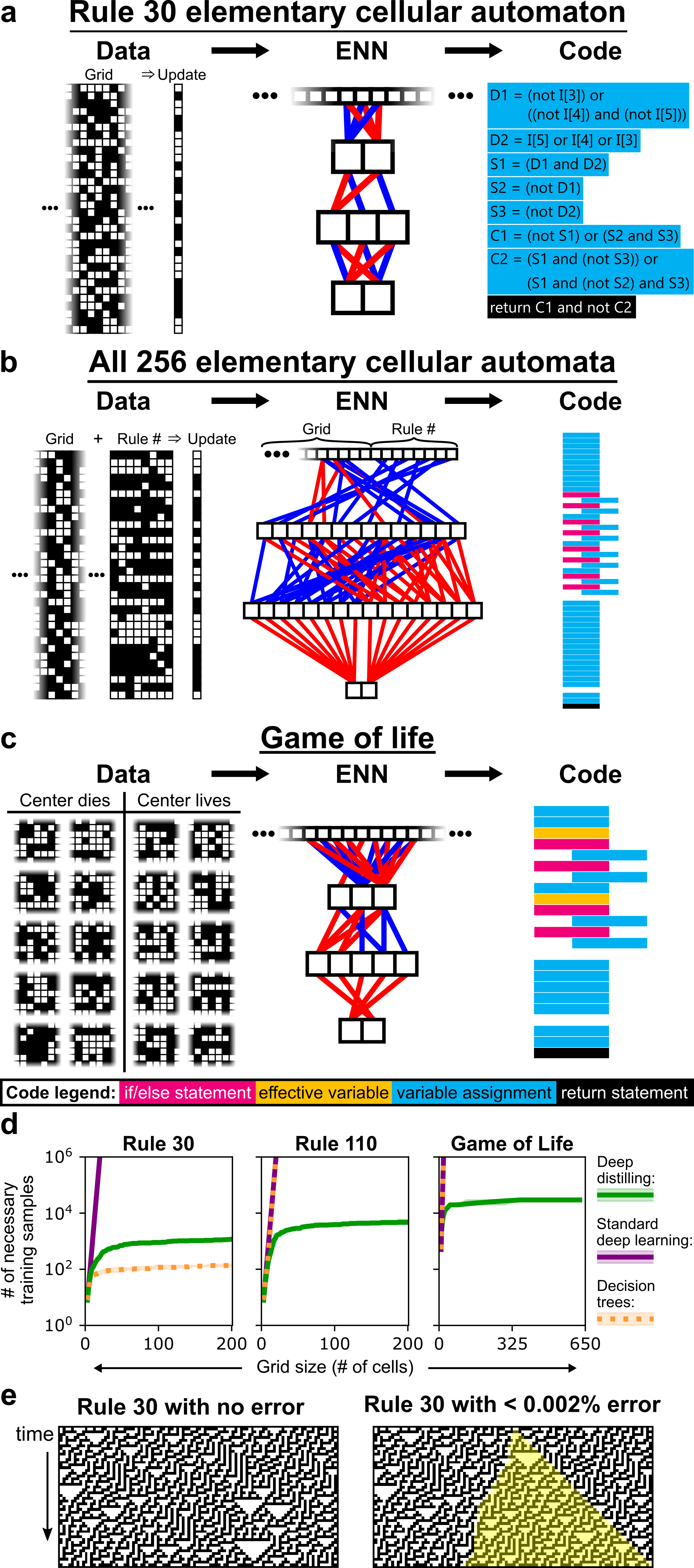}
  \end{minipage}\hfill
  \begin{minipage}[t]{0.25\textwidth}
    \vspace{0.01pt}
    \caption{
       \textbf{Deep distilling the rules of cellular automata from pictures.}Deep distilling was able to produce code that maps the current state of various CA to future states for \textbf{(a)} a single elementary CA (e.g. rule 30), \textbf{(b)} all 256 elementary CA simultaneously, and \textbf{(c)} the Game of life. The sign of ENN weights are shown in red (positive) and blue (negative), while magnitudes are not shown. Distilled code can be viewed in its entirety in Appendix A, while here it is shown simply as colored schematics (legend available). \textbf{(d)} Deep distilling can write this code even when trained only on a small fraction of the total possible grids, while other methods require almost all possible grids. \textbf{(e)} When a CA makes a single error, it spreads over time (highlighted in yellow).
    }
  \end{minipage}
\end{figure}

\subsection*{Distilling rules from cellular automata}

To test whether deep distilling can discover underlying rules from data, we applied it to cellular automata (CA). CA are grids consisting of cells in discrete states that evolve over time according to a set of rules, and as such have long been used in the physical, life, and computer sciences to model emergent phenomena. For each of the 256 elementary CA described by Wolfram (e.g. the chaotic rule 30 in Fig. 2a) \cite{wolframECA}, we randomly generated large one-dimensional grids and labeled each according to the center cell's state at the next time step. Deep distilling learned and wrote code that both perfectly reproduces the rule for each CA and correctly ignores all cells outside the central 3-cell neighborhood.
\par
In addition to learning individual CA, deep distilling also learned and wrote code for all 256 elementary CA at once. Training samples contained, in addition to the grid, an 8-bit string encoding a specific rule number (Fig. 2b). The distilled code, as expected, is longer and more complex than for individual CA (full code in Appendix A). Deep distilling discovered a simple algorithm that compares the grid's central neighborhood to a compressed, rule-specific truth table, an encoding different from standard descriptions of elementary CA rules. Interestingly, deep distilling was able to learn this encoding when trained on only 16 of the 256 CA rules. The symbolic nature of the algorithms discovered by deep distilling allows them to make these out-of-distribution generalizations to unencountered rule numbers.
\par
We also applied deep distilling to randomly generated two-dimensional grids from the Game of Life (Fig. 2c), the most famous CA due to its interesting emergent properties and Turing completeness \cite{gameoflife, turing_gol}. The distilled code perfectly describes the Game of Life rules (Appendix A), appropriately creating a distilled variable representing the total number of live cells in the central cell's neighborhood.
\par
While these CA have relatively simple underlying rules, they are non-trivial problems for other machine learning systems. For example, standard deep learning had difficulty with both elementary CA and the Game of Life when the grid became larger. It required essentially all $2^n$ possible $n$-cell grids to guarantee zero prediction error (Fig. 2d), meaning in practice it could only learn from small grids with complete training sets. When CA are simulated using a trained model without this guarantee of zero error, mistakes arise and propagate over time, destroying the accuracy of the results in a hard-to-detect way (Fig. 2e). Deep distilling, on the other hand, generates algorithms which inherently guarantee the absence of rare, unforeseen behaviors. Simpler explainable models such as decision trees were able to efficiently learn some of the elementary CA (e.g. rule 30). However, for 102 of the 256 rules (e.g. the Turing complete rule 110), decision tree learning failed because of ambiguity in choosing which feature to split at various nodes in the tree; this led to excessively large decision trees and the need to train on all $2^n$ sample grids to guarantee correct performance (Fig. 2d). Both standard deep learning and decision trees failed to generalize on the Game of Life (Fig. 2d).

\subsection*{Deep distilling algorithms that generalize and scale}

\begin{figure}[h!]
    \includegraphics[width=\textwidth]{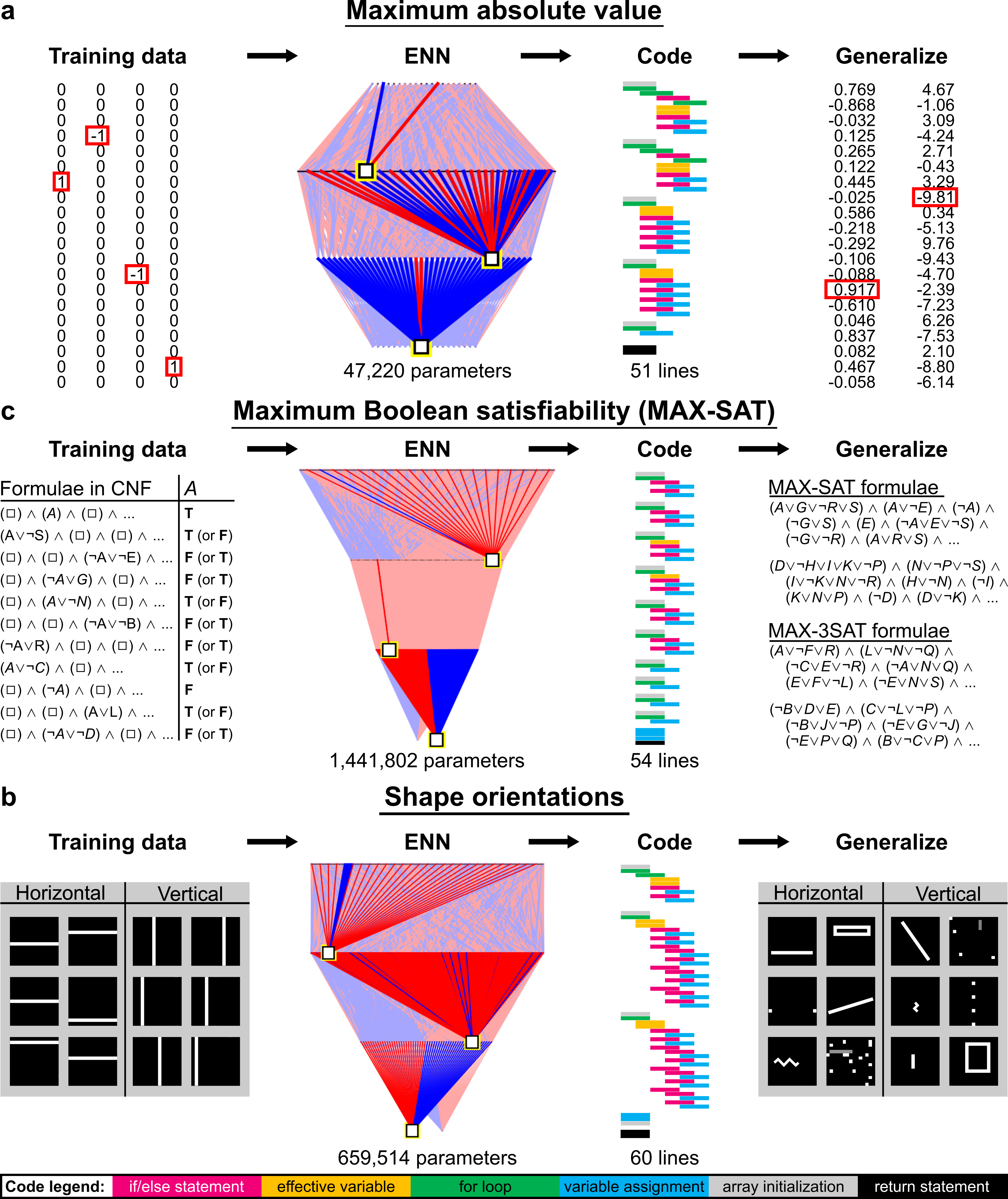}
    \caption{\textbf{Deep distilling learns generalizable code} Deep distilling discovered algorithms written in code \textbf{(a)} to find the maximum absolute value in an array, \textbf{(b)} to implement a greedy heuristic for the MAX-SAT problem, and \textbf{(c)} to determine the orientation of a shape in an image. In each case simple training samples were given, from which deep distilling wrote code that could generalize out-of-distribution for larger samples and for more complex cases. Schematic summaries of the code are shown here, with full code in Appendix A. In each layer of each ENN illustration is highlighted a random neuron with its incoming connections in order to demonstrate the magnitude and complexity of the trained ENNs.} 
    \label{fig3}
\end{figure}

While the difficulty of distilling CA rules increases with increased grid size, the complexity of the distilled rules does not. We next tested if deep distilling could discover algorithms whose rule-complexity scales with the size of the inputs. For example, deep distilling was able to learn and write code from a training set labeled by $\mathrm{argmax}_{x \in X} | x | $. Using only simple training samples containing just a single non-zero number, deep distilling discovered an algorithm that generalizes for any $X \in \mathbbm{R}^d$ for any $d=2,3,...$ (Fig. 3a, Fig. 4, Appendix A). The use of for-loops allowed the condensed code to be much smaller in size than the ENN. Usually the argmax function is computed by iterating once through the numbers and using a rewriteable variable to hold the largest number yet seen, but because the basic ENN architecture has no recurrent connections, the condensed code's variables are not rewriteable. Instead the distilled code compares each number to all the other numbers and their negatives, and then it chooses the number that won all of its comparisons.
\begin{figure}[h!]
    \centering
    \includegraphics[width=\textwidth]{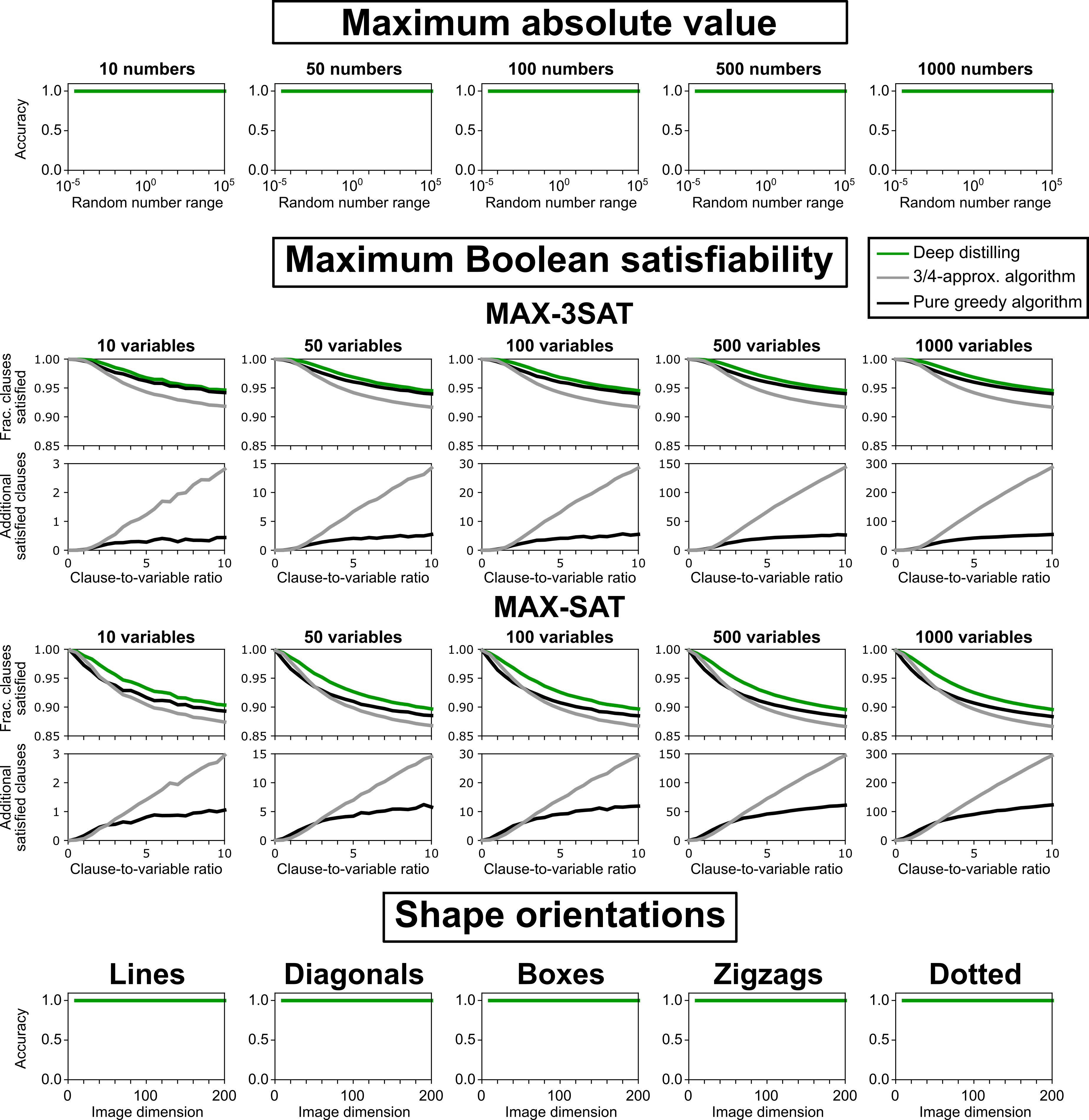}
    \raggedright
    \caption{\textbf{Deep distilling produces code that generalizes to arbitrary input sizes.} The size-generalized code distilled for MAX-3SAT and MAX-SAT were trained on data using only 8, 9, and 10 variables and 98, 99, and 100 clauses. The distilled code was able to perform well on Boolean formula of much larger sizes, even to 1000 variables and 10,000. For MAX-3SAT and MAX-SAT, the upper plots show the percentage of clauses that were satisfied as a function of the number of clauses by the distilled code, by the pure greedy algorithm, and by the 3/4-approximation algorithm. The lower plots show the absolute difference in clauses satisfied by the two human-designed algorithms compared to the distilled code (a positive difference indicates the distilled code satisfied more clauses). }
    \label{figs2}
\end{figure}
\par
We next considered a problem for which the optimal solution is presumably intractable (i.e. requires exponential run-time). One such task is the classic NP-hard maximum Boolean satisfiability (MAX-SAT) problem (Fig. 3c) \cite{karp_np}. Training on simple formulae with mostly empty clauses and no more than two literals per clause (Methods), deep distilling discovered a greedy algorithm very similar to a human-designed greedy algorithm with the best-known approximation ratio of 3/4 \cite{approximationSAT}. Both algorithms, when deciding to assign either TRUE or FALSE to a given variable in a Boolean formula, consider how many clauses will be satisfied as well as how many clauses will be left unsatisfiable by either assignment (Appendix A). However, while the 3/4-approximation algorithm weights these two considerations equally, the distilled algorithm learned to put greater weight on not leaving clauses unsatisfied. This difference allowed the distilled algorithm to outperform the 3/4-approximation algorithm on random MAX-3SAT Boolean formulae (where each clause contains three literals) and general MAX-SAT formulae, even for cases with orders of magnitude more clauses and variables than the training samples (Fig. 4). It also outperformed the pure greedy algorithm, which assigns TRUE or FALSE based on which satisfies the most clauses).
\par
We next considered a problem with ambiguous test cases, specifically the classification of a shape's orientation as vertical or horizontal in an image (Fig. 3b). Using only a few training images containing a single full-length white line, deep distilling learned a scalable algorithm that generalizes well in ambiguous cases (Appendix A). The distilled algorithm first compares the total brightness between rows and columns. Then it determines which rows or columns win more of these comparisons compared to the background, with closer cases resolved using nested logic statements. This novel algorithm offers certain benefits over standard human-designed algorithms such as convolutional filters and the Hough transform. For example, convolutional filters perform worse than the distilled algorithm with sparsely dotted lines, and the Hough transform performs worse on images with a low signal-to-noise ratio (Fig. 5). The distilled code furthermore provides a consistent standard for orientation classification even for ambiguous cases, such as zigzag and diagonal lines and patterns with multiple lines or background gradients (Figs. 4 and 6).
\begin{figure}[t!]
    \centering
    \includegraphics[width=\textwidth]{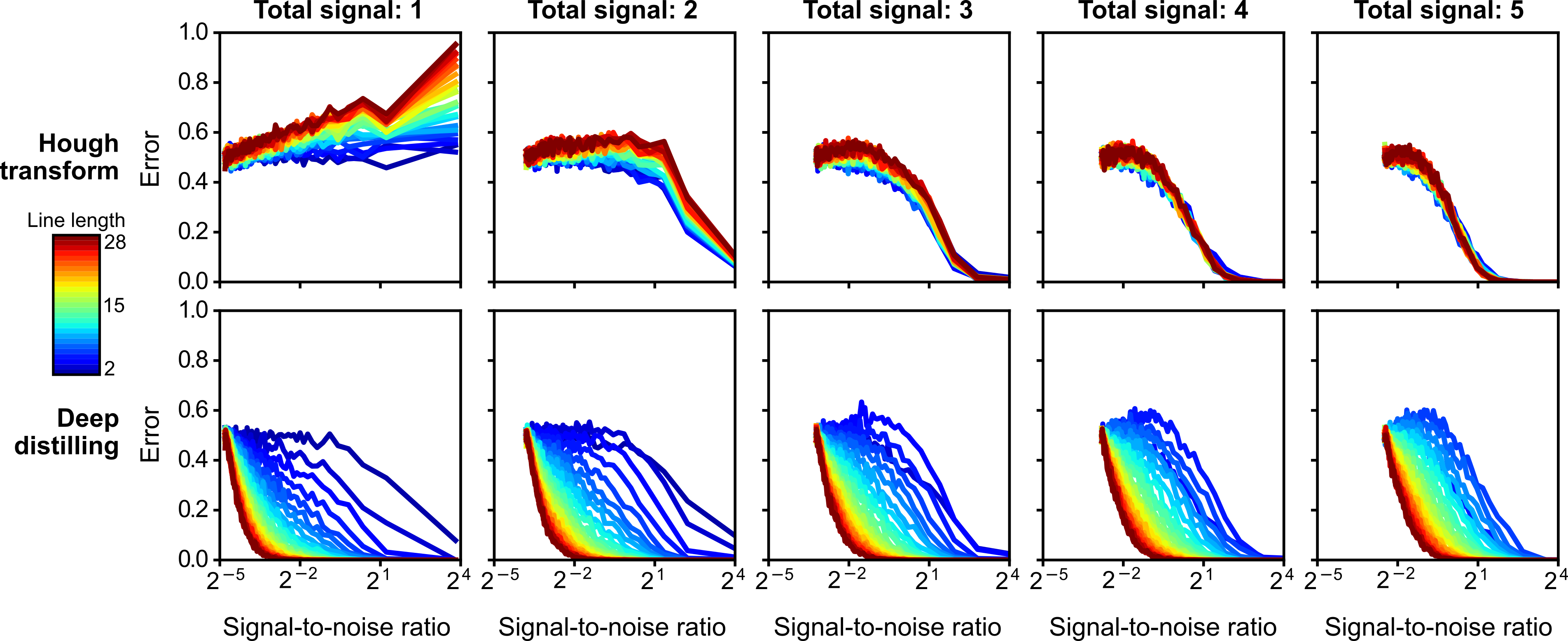}
    \raggedright
    \caption{\textbf{Deep distilling is more robust to images of low signal-to-noise ratio} Images were created with vertical and horizontal lines of different lengths, with the pixel value of each pixel in the line such that the sum total signal for the line was a fixed amount. Speckle noise was added to the images in various amounts for different signal-to-noise ratios. Results shown are the average error of each method when tested on 10,000 randomly generated images for each combination of signal and noise.}
    \label{figs1}
\end{figure}

\begin{figure}[b!]
  \begin{minipage}[t]{0.4\textwidth}
    \vspace{0.01pt}
    \caption{
       \textbf{Deep distilling produces an algorithm that assigns meaning in ambiguous cases.} Each of these images have ambiguously oriented shapes or gradients. Each image is labeled according to how the distilled code chooses to break this ambiguity and make a decision.
    }
  \end{minipage}\hfill
  \begin{minipage}[t]{0.55\textwidth}
    \vspace{0.01pt}
    \includegraphics[width=\textwidth]{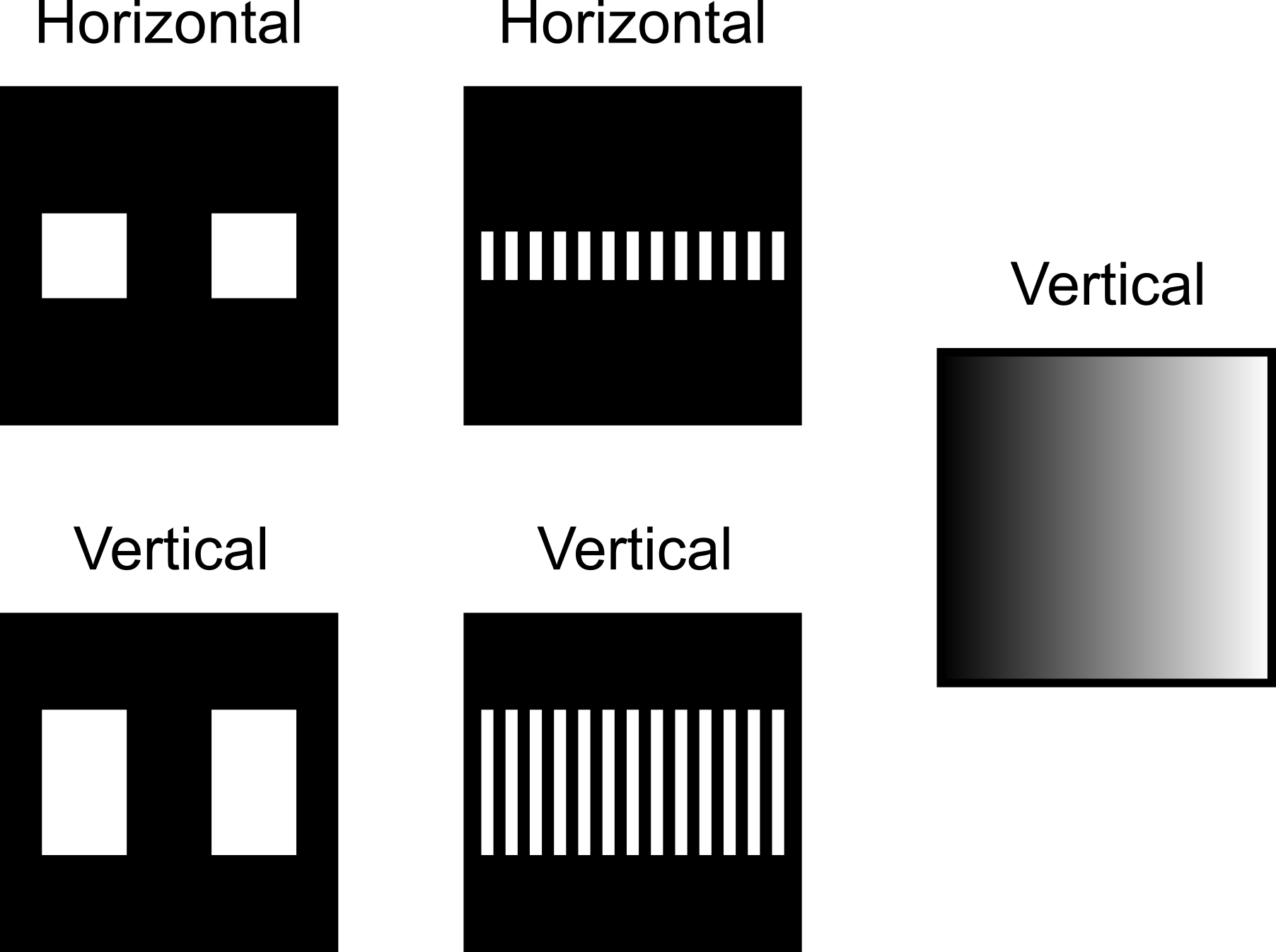}
  \end{minipage}
  
\end{figure}

\section*{Discussion}
Inductive programming (IP) has previously mostly focused on either very simple inductive problems or on using human-derived formal logic programming frameworks. Because deep distilling does not use search-based methods to write code, it works on a wide variety of problem types and can handle large problems that require non-trivial algorithms. However, while we have focused on deep distilling as a new approach for IP in particular, it could also be used for more general applications of automatic programming, such as generating code from descriptions. Others have used massive databases of coding samples from many programming languages (e.g. CodeNet \cite{codenet}) to train NLP models that mimic code-writing without truly understanding algorithm design. Instead, deep distilling could be combined with NLP to process the user's input and then design instructive training samples for deep distilling. One benefit of deep distilling is that it does not need to relearn how to code in new programming languages from scratch, but simply needs to be configured to translate the distilled algorithm into any desired language.
\par
The code written by deep distilling is complex enough to contain for-loops, distilled variables, nested logic statements, and several steps of information integration. It currently does not write code containing recursion, rewriteable variables, or functions beyond Boolean logic. However, it is straightforward to expand the complexity and depth of ENN architectures, such as previously described \cite{enns}, because the core principles by which ENNs are trained and condensed are general enough to support such extensions.
\par
Deep distilling may further provide new insights and inspiration for the theory of algorithm design. Because ENNs are designed to learn useful distinctions and to integrate them, they are able to discover algorithms how humans do, that is step-by-step without brute-force search methods. As a result, deep distilling is able to produce algorithms that can outperform human-designed algorithms in certain cases. This seems to be because deep distilling does not have the same biases that humans have and focus on how to best distinguish different patterns. In this way deep distilling presents a new machine-based reasoning approach that is able to extract human-understandable rules and principles from large datasets, implementing automated programming to discover generative rules and invent generalizable algorithms. In principle, these properties could enable deep distilling to automate the discovery and refinement of scientific models from high-dimensional observational and experimental data that may otherwise overwhelm human reasoning.

\section*{Acknowledgments}
The authors acknowledge the UTSW HR/HI grant for funding this research.

\section*{Author contributions}
P.J.B. and M.M.L. designed research; P.J.B. and K.V. performed research and contributed new analytical tools; P.J.B. analyzed data; P.J.B. and M.M.L. wrote the paper.

\section*{Competing interests}
The authors have filed a provisional patent related to this work.

\section*{Data and code availability}
The code and data used used to train and condense ENNs via deep distilling will be made available upon publication.

\raggedright
\printbibliography
\justifying

\newpage
\section*{Appendix A: Code from deep distilling}
On the following pages is the code as produced by deep distilling. For each problem, we have included the code twice. On the left side is the raw code output from the ENN condenser. This code has certain values hard-coded into it. On the right is the code generalized for arbitrary input size. All examples shown are written in Python.
\par
In each case the variables are given fairly nondescript names. Variables that start with "D" correspond to differentiae neurons in the ENN. Variables that start with "S" correspond to subconcept neurons in the ENN. Variables that start with "C" correspond to the output concept neurons. Some blank lines have been added to facilitate alignment of the left and right columns of code.

\newgeometry{left=1in, right=1in, top=1in, bottom=1in}

\definecolor{dkgreen}{rgb}{0,0.6,0}
\definecolor{gray}{rgb}{0.5,0.5,0.5}
\definecolor{mauve}{rgb}{0.58,0,0.82}

\lstset{frame=tb,
  language=Python,
  aboveskip=3mm,
  belowskip=3mm,
  showstringspaces=false,
  columns=flexible,
  basicstyle={\small\ttfamily},
  numbers=none,
  numberstyle=\tiny\color{gray},
  keywordstyle=\color{blue},
  commentstyle=\color{dkgreen},
  stringstyle=\color{mauve},
  breaklines=true,
  breakatwhitespace=true,
  tabsize=3
}

\clearpage

\begin{center}
\textbf{Distilled code to update a Rule 30 cellular automaton}
\end{center}
\begin{minipage}[t]{0.48\textwidth}
\begin{lstlisting}
import numpy as np

def rule30_3(I):
    #I is a 3-cell grid, with cell 1 being the cell to update
    
	D1 = (not I[0]) or ((not I[1]) and (not I[2]))
	
	D2 = I[2] or I[1] or I[0]
	
	
	S1 = (D1 and D2)
	
	S2 = (not D1)
	
	S3 = (not D2)
	
	C1 = (not S1) or (S2 and S3)
	
	C2 = (S1 and (not S3)) or (S1 and (not S2) and S3)
	
	return C1 and not C2
\end{lstlisting}
\end{minipage}
\begin{minipage}[t]{0.48\textwidth}
\begin{lstlisting}
import numpy as np

def rule30(I, n):
    #I is an n-cell grid, with cell (n-1)/2 being the cell to update
    
    D1 = (not I[(n-1)/2 - 1]) or ((not I[(n-1)/2]) and (not I[(n-1)/2 + 1]))

    D2 = I[(n-1)/2 + 1] or I[(n-1)/2] or I[(n-1)/2 - 1]
	
    S1 = (D1 and D2)
	
    S2 = (not D1)
	
    S3 = (not D2)
	
    C1 = (not S1) or (S2 and S3)
	
    C2 = (S1 and (not S3)) or (S1 and (not S2) and S3)
	
    return C1 and not C2
\end{lstlisting}
\end{minipage}\hfill

\begin{center}
\textbf{Distilled code to update a Rule 110 cellular automaton}
\end{center}
\begin{minipage}[t]{0.48\textwidth}
\begin{lstlisting}
import numpy as np

def rule110_3(I):
    #I is a 3-cell grid, with cell 1 being the cell to update
    
    D1 = I[1] or I[2]
	
    D2 = (not I[0]) or (not I[1]) or (not I[2])
	
	
    S1 = (D1 and D2)
	
    S2 = (not D1)
	
    S3 = (not D2)
	
    C1 = (not S1) or (S2 and S3)
	
    C2 = (S1 and (not S3)) or (S1 and (not S2) and S3)
	
    return C1 and not C2
\end{lstlisting}
\end{minipage}
\begin{minipage}[t]{0.48\textwidth}
\begin{lstlisting}
import numpy as np

def rule110(I, n):
    #I is an n-cell grid, with cell (n-1)/2 being the cell to update
    
    D1 = I[(n-1)/2] or I[(n-1)/2 + 1]
	
    D2 = (not I[(n-1)/2 - 1]) or (not I[(n-1)/2]) or (not I[(n-1)/2 + 1])
	
    S1 = (D1 and D2)
	
    S2 = (not D1)
	
    S3 = (not D2)
	
    C1 = (not S1) or (S2 and S3)
	
    C2 = (S1 and (not S3)) or (S1 and (not S2) and S3)
	
    return C1 and not C2
\end{lstlisting}
\end{minipage}\hfill

\clearpage

\begin{center}
\textbf{Distilled code to update any of the 256 elementary cellular automata}
\end{center}
\lstset{basicstyle={\scriptsize\ttfamily}}
\begin{minipage}[t]{0.48\textwidth}
\begin{lstlisting}
import numpy as np

def elementary_automata_3(I1, I2):
    #I1 is the 8-bit encoding of the rule number. #I2 is a 3-cell grid, with cell 1 being the cell to update
    
    
	D1 = (not I1[0])
	D2 = (not I1[1])
	D3 = (not I1[2])
	D4 = (not I1[3])
	D5 = (not I1[4])
	D6 = (not I1[5])
	D7 = (not I1[6])
	D8 = (not I1[7])
	
	D9 = (not I2[1])
	D10 = I2[1]
	
	D11 = 0.5
	if ((not I2[0]) and (not I2[2])):
		D11 = 1
	elif (not I2[2]) or (not I2[(n-1)/2-):
		D11 = 0
	
	D12 = 0.5
	if (I2[0] and (not I2[2])):
		D12 = 1
	elif (not I2[2]) or I2[0]:
		D12 = 0
	
	D13 = 0.5
	if (I2[2] and (not I2[0])):
		D13 = 1
	elif (not I2[0]) or I2[2]:
		D13 = 0
	
	D14 = 0.5
	if (I2[0] and I2[2]):
		D14 = 1
	elif I2[2] or I2[0]:
		D14 = 0
	
	S1 = (D14 and D1 and D10)
	S2 = ((not D1) and (not D9) and (not D11))
	S3 = (D12 and D2 and D10)
	S4 = ((not D2) and (not D9) and (not D13))
	S5 = (D14 and D3 and D9)
	S6 = ((not D3) and (not D10) and (not D11))
	S7 = (D12 and D4 and D9)
	S8 = ((not D4) and (not D10) and (not D13))
	S9 = (D13 and D5 and D10)
	S10 = ((not D5) and (not D9) and (not D12))
	S11 = (D11 and D6 and D10)
	S12 = ((not D6) and (not D9) and (not D14))
	S13 = (D13 and D7 and D9)
	S14 = ((not D7) and (not D10) and (not D12))
	S15 = (D11 and D8 and D9)
	S16 = ((not D8) and (not D10) and (not D14))
	
	C1 = S15 or S13 or S11 or S9 or S7 or S5 or S3 or S1
	C2 = S16 or S14 or S12 or S10 or S8 or S6 or S4 or S2
	
	return C1 and not C2
\end{lstlisting}
\end{minipage}\hfill
\begin{minipage}[t]{0.48\textwidth}
\begin{lstlisting}
import numpy as np

def elementary_automata(I1, I2, n):
    #I1 is the 8-bit encoding of the rule number. #I2 is an n-cell grid, with cell (n-1)/2 being the cell to update
    
	D1 = (not I1[0])
	D2 = (not I1[1])
	D3 = (not I1[2])
	D4 = (not I1[3])
	D5 = (not I1[4])
	D6 = (not I1[5])
	D7 = (not I1[6])
	D8 = (not I1[7])
	
	D9 = (not I2[(n-1)/2])
	D10 = I2[(n-1)/2]
	
	D11 = 0.5
	if ((not I2[(n-1)/2 - 1]) and (not I2[(n-1)/2 + 1])):
		D11 = 1
	elif (not I2[(n-1)/2 + 1]) or (not I2[(n-1)/2 - 1]):
		D11 = 0
	
	D12 = 0.5
	if (I2[(n-1)/2 - 1] and (not I2[(n-1)/2 + 1])):
		D12 = 1
	elif (not I2[(n-1)/2 + 1]) or I2[(n-1)/2 - 1]:
		D12 = 0
	
	D13 = 0.5
	if (I2[(n-1)/2 + 1] and (not I2[(n-1)/2 - 1])):
		D13 = 1
	elif (not I2[(n-1)/2 - 1]) or I2[(n-1)/2 + 1]:
		D13 = 0
	
	D14 = 0.5
	if (I2[(n-1)/2 - 1] and I2[(n-1)/2 + 1]):
		D14 = 1
	elif I2[(n-1)/2 + 1] or I2[(n-1)/2 - 1]:
		D14 = 0
	
	S1 = (D14 and D1 and D10)
	S2 = ((not D1) and (not D9) and (not D11))
	S3 = (D12 and D2 and D10)
	S4 = ((not D2) and (not D9) and (not D13))
	S5 = (D14 and D3 and D9)
	S6 = ((not D3) and (not D10) and (not D11))
	S7 = (D12 and D4 and D9)
	S8 = ((not D4) and (not D10) and (not D13))
	S9 = (D13 and D5 and D10)
	S10 = ((not D5) and (not D9) and (not D12))
	S11 = (D11 and D6 and D10)
	S12 = ((not D6) and (not D9) and (not D14))
	S13 = (D13 and D7 and D9)
	S14 = ((not D7) and (not D10) and (not D12))
	S15 = (D11 and D8 and D9)
	S16 = ((not D8) and (not D10) and (not D14))
	
	C1 = S15 or S13 or S11 or S9 or S7 or S5 or S3 or S1
	C2 = S16 or S14 or S12 or S10 or S8 or S6 or S4 or S2
	
	return C1 and not C2
\end{lstlisting}
\end{minipage}\hfill
\clearpage

\begin{center}
\textbf{Distilled code to update a Game of Life cellular automaton}
\end{center}
\lstset{basicstyle={\small\ttfamily}}
\begin{minipage}[t]{0.48\textwidth}
\begin{lstlisting}
import numpy as np

def game_of_life_3(I):
    #I is a 3x3 grid, with the center cell being the cell to update

	D1 = I[1, 1]
	
	D2 = 0
	part_sum = (I[0,0] + I[0,1] + I[0,2] + I[1,0] + I[1,2] + I[2,0] + I[2,1] + I[2,2])
	
	
	
	if part_sum > 3:
		D2 = 1
	elif part_sum <= 3:
		D2 = -1
	
	D3 = 0
	part_sum = (I[0,0] + I[0,1] + I[0,2] + I[1,0] + I[1,2] + I[2,0] + I[2,1] + I[2,2])
	
	
	
	if part_sum > 1:
		D3 = 1
	elif part_sum <= 1:
		D3 = -1
	
	S1 = (not D2>0)
	
	S2 = (not D3>0)
	
	S3 = ((not D1>0) and (not D3>0))
	
	S4 = (D2>0 and D3>0)
	
	S5 = (D1>0 and D2>0 and D3>0)
	
	C1 = (S4 and S5)
	
	C2 = (S1 and S3) or (S1 and S2 and (not S3))
	
	return C1 and not C2
\end{lstlisting}
\end{minipage}\hfill
\begin{minipage}[t]{0.48\textwidth}
\begin{lstlisting}
import numpy as np

def game_of_life(I, n):
    #I is an nxn grid, with the center cell being the cell to update

	D1 = I[(n-1)/2, (n-1)/2]
	
	D2 = 0
	part_sum = (I[(n-1)/2-1, (n-1)/2-1] + I[(n-1)/2-1, (n-1)/2] + I[(n-1)/2-1, (n-1)/2+1] + I[(n-1)/2, (n-1)/2-1] + I[(n-1)/2, (n-1)/2+1] + I[(n-1)/2+1, (n-1)/2-1] + I[(n-1)/2+1, (n-1)/2] + I[(n-1)/2+1, (n-1)/2+1])
	if part_sum > 3:
		D2 = 1
	elif part_sum <= 3:
		D2 = -1
	
	D3 = 0
	part_sum = (I[(n-1)/2-1, (n-1)/2-1] + I[(n-1)/2-1, (n-1)/2] + I[(n-1)/2-1, (n-1)/2+1] + I[(n-1)/2, (n-1)/2-1] + I[(n-1)/2, (n-1)/2+1] + I[2, (n-1)/2-1] + I[(n-1)/2+1, (n-1)/2] + I[(n-1)/2+1, (n-1)/2+1])
	if part_sum > 1:
		D3 = 1
	elif part_sum <= 1:
		D3 = -1
	
	S1 = (not D2>0)
	
	S2 = (not D3>0)
	
	S3 = ((not D1>0) and (not D3>0))
	
	S4 = (D2>0 and D3>0)
	
	S5 = (D1>0 and D2>0 and D3>0)
	
	C1 = (S4 and S5)
	
	C2 = (S1 and S3) or (S1 and S2 and (not S3))
	
	return C1 and not C2
\end{lstlisting}
\end{minipage}\hfill
\clearpage

\begin{center}
\textbf{Distilled code to find the maximum absolute value}
\end{center}
\lstset{basicstyle={\footnotesize\ttfamily}}
\begin{minipage}[t]{0.48\textwidth}
\begin{lstlisting}
import numpy as np
import random

def absmax_20(I):
    #I is an array of 20 numbers

	D1 = np.zeros((20, 20))
	for i in range(20):
		for j in range(20):
			if i == j:
				continue
			value_1 = I[i]
			value_2 = I[j]
			if value_1 > value_2:
				D1[i,j] = 1
			elif value_1 < value_2:
				D1[i,j] = -1
			
	D2 = np.zeros((20, 20))
	for i in range(20):
		for j in range(20):
			if i == j:
				continue
			value_1 = I[i]
			value_2 = I[j]
			if value_1 > -value_2:
				D2[i,j] = 1
			elif value_1 < -value_2:
				D2[i,j] = -1

	S1 = np.zeros(20)
	for i in range(20):
		row_sum_1 = np.sum(D1[i, :])
		row_sum_2 = np.sum(D2[i, :])
		if row_sum_1 < 18:
			S1[i] = -1
		elif row_sum_2 < 18:
			S1[i] = -1
		elif row_sum_1 + row_sum_2 > -37:
			S1[i] = 1
		else:
			S1[i] = -1
		
	S2 = np.zeros(20)
	for i in range(20):
		row_sum_1 = np.sum(D1[i, :])
		row_sum_2 = np.sum(D2[i, :])
		if row_sum_1 > -18:
			S2[i] = -1
		elif row_sum_2 > -18:
			S2[i] = -1
		elif -row_sum_1 - row_sum_2 > -37:
			S2[i] = 1
		else:
			S2[i] = -1

	C = np.zeros(20)
	for i in range(20):
		C[i] = 20*S2[i] + 20*S1[i] - np.sum(S2) - np.sum(S1)

	results = np.where(C==max(C))[0]
	return random.choice(results)
\end{lstlisting}
\end{minipage}\hfill
\begin{minipage}[t]{0.48\textwidth}
\begin{lstlisting}
import numpy as np
import random

def absmax(I, n):
    #I is an array of n numbers

	D1 = np.zeros((n, n))
	for i in range(n):
		for j in range(n):
			if i == j:
				continue
			value_1 = I[i]
			value_2 = I[j]
			if value_1 > value_2:
				D1[i,j] = 1
			elif value_1 < value_2:
				D1[i,j] = -1
			
	D2 = np.zeros((n, n))
	for i in range(n):
		for j in range(n):
			if i == j:
				continue
			value_1 = I[i]
			value_2 = I[j]
			if value_1 > -value_2:
				D2[i,j] = 1
			elif value_1 < -value_2:
				D2[i,j] = -1

	S1 = np.zeros(n)
	for i in range(n):
		row_sum_1 = np.sum(D1[i, :])
		row_sum_2 = np.sum(D2[i, :])
		if row_sum_1 < n-2:
			S1[i] = -1
		elif row_sum_2 < n-2:
			S1[i] = -1
		elif row_sum_1 + row_sum_2 > 3-2*n:
			S1[i] = 1
		else:
			S1[i] = -1
		
	S2 = np.zeros(n)
	for i in range(n):
		row_sum_1 = np.sum(D1[i, :])
		row_sum_2 = np.sum(D2[i, :])
		if row_sum_1 > 2-n:
			S2[i] = -1
		elif row_sum_2 > 2-n:
			S2[i] = -1
		elif -row_sum_1 - row_sum_2 > 3-2*n:
			S2[i] = 1
		else:
			S2[i] = -1

	C = np.zeros(n)
	for i in range(n):
		C[i] = n*S2[i] + n*S1[i] - np.sum(S2) - np.sum(S1)

	results = np.where(C==max(C))[0]
	return random.choice(results)
\end{lstlisting}
\end{minipage}\hfill
\clearpage

\begin{center}
\textbf{Distilled code to find a shape's orientation}
\end{center}
\lstset{basicstyle={\scriptsize\ttfamily}}
\begin{minipage}[t]{0.48\textwidth}
\begin{lstlisting}
import numpy as np
import random

def orientation_28(I, n):
    #I is an input image that is 28x28

	D = np.zeros((28, 28))
	for i in range(28):
		for j in range(28):
			col_sum = np.sum(I[:, i])
			row_sum = np.sum(I[j, :])
			if col_sum > row_sum:
				D[i,j] = 1
			elif col_sum < row_sum:
				D[i,j] = -1

	S1 = np.zeros(28)
	for i in range(28):
		row_sum = np.sum(D[i, :])
		offrow_sum = (np.sum(D) - np.sum(D[i, :]))
		if offrow_sum < -29:
			S1[i] = 1
		elif offrow_sum > -27:
			S1[i] = -1
		elif offrow_sum == -27:
			if np.all(D[i, :]==1):
				S1[i] = 1
			elif not np.all(D[i, :]==1):
				S1[i] = -1
		elif offrow_sum == -28:
			if row_sum > 0:
				S1[i] = 1
			elif row_sum < 0:
				S1[i] = -1
		elif offrow_sum == -29:
			if not np.all(D[i, :]==-1):
				S1[i] = 1
			elif np.all(D[i, :]==-1):
				S1[i] = -1
		
	S2 = np.zeros(28)
	for i in range(28):
		offcol_sum = (np.sum(D) - np.sum(D[:, i]))
		col_sum = np.sum(D[:, i])
		if offcol_sum < 27:
			S2[i] = -1
		elif offcol_sum > 29:
			S2[i] = 1
		elif offcol_sum == 29:
			if np.all(D[:, i]==1):
				S2[i] = -1
			elif not np.all(D[:, i]==1):
				S2[i] = 1
		elif offcol_sum == 28:
			if col_sum > 0:
				S2[i] = -1
			elif col_sum < 0:
				S2[i] = 1
		elif offcol_sum == 27:
			if not np.all(D[:, i]==-1):
				S2[i] = -1
			elif np.all(D[:, i]==-1):
				S2[i] = 1
		
	C1 = np.sum(S1) - np.sum(S2)
	C2 = np.sum(S2) - np.sum(S1)
	C = [C1, C2]

	results = np.where(C==max(C))[0]
	return random.choice(results)
\end{lstlisting}
\end{minipage}\hfill
\begin{minipage}[t]{0.48\textwidth}
\begin{lstlisting}
import numpy as np
import random

def orientation(I, n):
    #I is an input image that is nxn

	D = np.zeros((n, n))
	for i in range(n):
		for j in range(n):
			col_sum = np.sum(I[:, i])
			row_sum = np.sum(I[j, :])
			if col_sum > row_sum:
				D[i,j] = 1
			elif col_sum < row_sum:
				D[i,j] = -1

	S1 = np.zeros(n)
	for i in range(n):
		row_sum = np.sum(D[i, :])
		offrow_sum = (np.sum(D) - np.sum(D[i, :]))
		if offrow_sum < -1-n:
			S1[i] = 1
		elif offrow_sum > 1-n:
			S1[i] = -1
		elif offrow_sum == 1-n:
			if np.all(D[i, :]==1):
				S1[i] = 1
			elif not np.all(D[i, :]==1):
				S1[i] = -1
		elif offrow_sum == -n:
			if row_sum > 0:
				S1[i] = 1
			elif row_sum < 0:
				S1[i] = -1
		elif offrow_sum == -1-n:
			if not np.all(D[i, :]==-1):
				S1[i] = 1
			elif np.all(D[i, :]==-1):
				S1[i] = -1
		
	S2 = np.zeros(n)
	for i in range(n):
		offcol_sum = (np.sum(D) - np.sum(D[:, i]))
		col_sum = np.sum(D[:, i])
		if offcol_sum < n-1:
			S2[i] = -1
		elif offcol_sum > n+1:
			S2[i] = 1
		elif offcol_sum == n+1:
			if np.all(D[:, i]==1):
				S2[i] = -1
			elif not np.all(D[:, i]==1):
				S2[i] = 1
		elif offcol_sum == n:
			if col_sum > 0:
				S2[i] = -1
			elif col_sum < 0:
				S2[i] = 1
		elif offcol_sum == n-1:
			if not np.all(D[:, i]==-1):
				S2[i] = -1
			elif np.all(D[:, i]==-1):
				S2[i] = 1
		
	C1 = np.sum(S1) - np.sum(S2)
	C2 = np.sum(S2) - np.sum(S1)
	C = [C1, C2]

	results = np.where(C==max(C))[0]
	return random.choice(results)
\end{lstlisting}
\end{minipage}\hfill
\clearpage

\begin{center}
\textbf{Distilled code to find the best assignment for MAX-SAT}
\end{center}
\lstset{basicstyle={\scriptsize\ttfamily}}
\begin{minipage}[t]{0.48\textwidth}
\begin{lstlisting}
import numpy as np

def maxsat_10_50(I):
    #I is an input of size 10x50 (5 one-hot-encoded Boolean variables, 50 clauses)

	D1 = np.zeros(50)
	for i in range(50):
		if np.any(I[2:, i]!=0):
			D1[i] = -1
		else:
			D1[i] = 1

	D2 = np.zeros(50)
	for i in range(50):
		if np.any(I[2:, i]!=0):
			D2[i] = 1
		else:
			D2[i] = -1

	D3 = np.zeros(50)
	for i in range(50):
		col_mean = np.mean(I[2:, i])
		if I[1, i] + col_mean - I[0, i] > 0:
			D3[i] = 1
		else:
			D3[i] = -1
		
	D4 = np.zeros(50)
	for i in range(50):
		col_mean = np.mean(I[2:, i])
		if I[0, i] + col_mean - I[1, i] > 0:
			D4[i] = 1
		else:
			D4[i] = -1
		
	D5 = np.zeros(50)
	for i in range(50):
		if (I[0, i] and (not I[1, 0])):
			D5[i] = 1
		elif (not I[1, 0]) or I[0, i]:
			D5[i] = -1
		
	D6 = np.zeros(50)
	for i in range(50):
		if (I[1, i] and (not I[0, 0])):
			D6[i] = 1
		elif (not I[0, 0]) or I[1, i]:
			D6[i] = -1

	S1 = np.zeros(50)
	for i in range(50):
		S1[i] = (D6[i]>0 and D1[i]>0)
		
	S2 = np.zeros(50)
	for i in range(50):
		S2[i] = (D2[i]>0 and D3[i]>0)
		
	S3 = np.zeros(50)
	for i in range(50):
		S3[i] = (D5[i]>0 and D1[i]>0)
		
	S4 = np.zeros(50)
	for i in range(50):
		S4[i] = (D2[i]>0 and D4[i]>0)
		
	C1 = 10.0*np.sum(S3) + 2.298*np.sum(S4) - 2.298*np.sum(S2) - 10.0*np.sum(S1)
	C2 = 10.0*np.sum(S1) + 2.298*np.sum(S2) - 2.298*np.sum(S4) - 10.0*np.sum(S3)
	C = [C1, C2]

	return np.exp(C)/np.sum(np.exp(C))
\end{lstlisting}
\end{minipage}\hfill
\begin{minipage}[t]{0.48\textwidth}
\begin{lstlisting}
import numpy as np

def maxsat(I, n, m):
    #I is an input of size mxn (m one-hot-encoded Boolean variables, n clauses)

	D1 = np.zeros(n)
	for i in range(n):
		if np.any(I[2:, i]!=0):
			D1[i] = -1
		else:
			D1[i] = 1

	D2 = np.zeros(n)
	for i in range(n):
		if np.any(I[2:, i]!=0):
			D2[i] = 1
		else:
			D2[i] = -1

	D3 = np.zeros(n)
	for i in range(n):
		col_mean = np.mean(I[2:, i])
		if I[1, i] + col_mean - I[0, i] > 0:
			D3[i] = 1
		else:
			D3[i] = -1
		
	D4 = np.zeros(n)
	for i in range(n):
		col_mean = np.mean(I[2:, i])
		if I[0, i] + col_mean - I[1, i] > 0:
			D4[i] = 1
		else:
			D4[i] = -1
		
	D5 = np.zeros(n)
	for i in range(n):
		if (I[0, i] and (not I[1, 0])):
			D5[i] = 1
		elif (not I[1, 0]) or I[0, i]:
			D5[i] = -1
		
	D6 = np.zeros(n)
	for i in range(n):
		if (I[1, i] and (not I[0, 0])):
			D6[i] = 1
		elif (not I[0, 0]) or I[1, i]:
			D6[i] = -1

	S1 = np.zeros(n)
	for i in range(n):
		S1[i] = (D6[i]>0 and D1[i]>0)
		
	S2 = np.zeros(n)
	for i in range(n):
		S2[i] = (D2[i]>0 and D3[i]>0)
		
	S3 = np.zeros(n)
	for i in range(n):
		S3[i] = (D5[i]>0 and D1[i]>0)
		
	S4 = np.zeros(n)
	for i in range(n):
		S4[i] = (D2[i]>0 and D4[i]>0)
		
	C1 = 10.0*np.sum(S3) + 2.298*np.sum(S4) - 2.298*np.sum(S2) - 10.0*np.sum(S1)
	C2 = 10.0*np.sum(S1) + 2.298*np.sum(S2) - 2.298*np.sum(S4) - 10.0*np.sum(S3)
	C = [C1, C2]

	return np.exp(C)/np.sum(np.exp(C))
\end{lstlisting}
\end{minipage}\hfill

\newpage
\newgeometry{left=1.5in, right=1.5in, top=1in, bottom=1in}
\section*{Appendix B: Methods}
Deep distilling produces code meant to map inputs to outputs given training data. It is a two-step process, first training ENNs for explainable deep learning, and then condensing the ENN to readable code.
\subsection*{ENN training}
All ENNs trained here use the basic ENN architecture and training methods described previously \cite{enns}. Briefly, basic ENN training first divides up all of the training samples into \textit{subconcepts}, which are subsets of similar training samples with the same target output (i.e. the same target \textit{concept}). A first hidden layer of neurons (called \textit{differentia neurons}) such that each neuron distinguishes a pair of subconcepts and is designed by computing linear support-vector machines (SVMs) between these subconcepts, giving the learned weights and bias factor to that differentia neuron. A second hidden layer of neurons (called \textit{subconcept neurons}) is constructed such that each neuron represents a specific subconcept and is designed by computing SVMs---using the differentia neuron outputs---between the training samples of that subconcept \textit{versus} all other training samples. A final output layer of neurons (called \textit{concept neurons}) is constructed such that each neuron represents a specific concept and is designed by computing SVMs---using the subconcept neuron outputs---between the training samples of that concept \textit{versus} all other training samples. Each neuron's output is given by $y=\mathrm{sign}(\mathbf{w} \cdot \mathbf{x} + b)$, where $\mathbf{w} \in \mathbb{R}^m$ is a set of weights applied to the neuron's incoming inputs $\mathbf{x} \in \mathbb{R}^m$, and $b \in \mathbb{R}$ is the neuron's bias factor.
\par
The one alteration we now introduce is two new methods for learning subconcepts within the training data, which we developed so that ENN training automatically learns an appropriate number of subconcepts. Previously, the training samples belonging to each concept were divided into subconcepts using hierarchical clustering, finding a fixed value to cut each hierarchical tree such that the total number of cut clusters across all concepts was equal to a user-defined number of clusters (i.e. subconcepts). Before implementing either method, for those tasks for which the inputs were discretely symbolic, with a 1 indicating the presence of a feature and 0 representing its absence, we first divided the concepts into new concepts such that each training sample shared at least one feature with another training sample. This was done to make sure that each concept had a shared familial resemblance. Representing each concept as a graph with nodes representing each sample and the presence of an edge representing shared features, depth-first search finds isolated graphs (i.e. components) that ultimately form the new concepts.
\par
To determine how much to further divide the concepts, we used one of two methods. One way used hierarchical clustering as usual, but instead of choosing the cutoff value for the trees such that it resulted in a predefined number of subconcepts, the value was chosen so that the minimum number of disjoint (i.e. linearly separable) subconcept regions were formed. First, hierarchical clustering was performed on the training samples from each concept separately, and all the hierarchical trees were cut at whatever cutoff value that would result in a predefined minimum number of subconcepts. Linear support vector machines (SVMs) were computed for every pair of subconcepts to find hyperplanes that separate the subsconcepts' training samples. This hyperplane divides up input space into two half-spaces, $\textbf{w} \cdot \textbf{x}^{(-)} + b < 0$ and $\textbf{w} \cdot \textbf{x}^{(+)} + b > 0$, with all the negative half-space points $\textbf{x}^{(-)}$ satisfying the first inequality and the positive half-space points $\textbf{x}^{(+)}$ satisfying the second inequality. If the SVM hyperplane did not perfectly separate the two subconcept's training samples into separate half-spaces, then the subconcepts were not linearly separable (or at least not easily separable enough). If any of the pairs of subconcepts were not linearly separable, then the desired number of subconcepts was increased and the hierarchical trees were re-cut at a new value to yield an additional subconcept. This process was repeated until all subconcepts were linearly separated from one another.
\par
The other method for dividing subconcepts checked the concepts for linear separability in the same manner as above. However, this time the misclassification error of each SVM was also computed and stored. For the pair of subconcepts with the greatest misclassification error, the misclassified training samples from one subconcept were removed and placed into a new subconcept. This process was repeated until all of the subconcepts were linearly separated from one another.

\subsection*{ENN condenser}
Interpretation of the ENN as computer code was performed by our ENN condenser, a meta-program that receives as input a trained ENN and writes computer code that performs the exact step-by-step reasoning process. This code can be written in any desired programming language (or even prose), and we have chosen here to illustrate it in Python. The ENN condenser can be broken down into two modules---organizing groups of neurons and then interpreting them---for any possible ENN architecture, although here we only demonstrate it for the basic ENN architecture.
\par
\subsubsection*{Organizing ENN neurons}
Organization of neurons involves separating them into separate groups and then arranging them within the group in a structure such that the neurons can be iterated over (e.g. a vector or matrix). The input neurons are pre-arranged by the user, for example as a vector or in a grid (e.g. images). To arrange neurons into groups, the connectivity pattern of each neuron is determined, and then neurons are placed into groups with related connectivity.
\par
To determine the connectivity pattern for each neuron, its incoming weights are normalized, dividing by the absolute value of the lowest magnitude non-zero weight. If any of these normalized weights $\mathbf{w}_\text{norm}$ are not close to an integer value (i.e. $|w_\text{norm}-\mathrm{round}(w_\text{norm})| > \epsilon$ for some $\epsilon>0$), then a number $\alpha>1$ is found such that when all of the weights were multiplied by $\alpha$ they satisfy $|\alpha w_\text{norm}-\mathrm{round}(\alpha w_\text{norm})| < \epsilon$. The finalized weights of the neuron are $\mathbf{w}=\mathrm{round}(\alpha \mathbf{w_\text{norm}})$.
\par
For each neuron, the unique non-zero values of $\mathbf{w}$ are represented in a vector $\mathbf{u} \in \mathbb{
Z}^{m'}$ with $m' \leq m$, and the neuron's connectivity patterns are examined. For each element $u_k$ of $\mathbf{u}$, some subset of the incoming connections are weighted by $u_k$. The connectivity pattern of these synapses we denote by $p^{(k)} (P, \mathbf{g},\mathbf{d})$, a vector containing the indices of all the incoming neuron connections that are weighted by $u_k$. $P$ represents a particular class of connectivity patterns ($p^{(k)} \in P$), $\mathbf{g}$ represents the indices of the neuron groups to which $P$ is applied, and $\mathbf{d}$ represents the indices that define which exact pattern applies for $u_k$.  We can then define $c_k( \mathbf{x}; p^{(k)} ) = \sum_{i \in p^{(k)}}{x_i}$ to serve as a reduction function that condenses all the inputs from $\mathbf{x}$ that are in $p^{(k)}$ to a single value $c_k$, which we have termed a \textit{condensed variable}. This allows the neuron's output to be rewritten as $y=\mathrm{sign} (b + \sum_k {u_k c_k})$.
\par
Examples of classes of connectivity patterns $P$ include: a column in a matrix, a row in a matrix, multiple columns or rows in a matrix, all elements in a vector, a single element in a matrix, etc. The indices $\mathbf{d}$ will specify an exact connectivity pattern in one of these classes, for example indicating the exact row $\mathbf{d}=(i)$ from a matrix or the exact indices $(i,j)$ of the element in a matrix. For example, the connectivity pattern of each neuron $i$ in Figure 1b's single index example would be $p^{(1)} ( \text{matrix-row}, 1, i)$, since all the weights come from row $i$ of the 1st neuron group. The neurons in the nested indices example have two unique weights, and for each neuron $(i,j)$ the connectivity patterns are $p^{(1)} ( \text{vector-element}, 1, j )$ and $p^{(2)} (\text{vector-element}, 2, i )$ for the two elements of $\mathbf{u}=(w_1,w_2)$, respectively. 
\par
Neurons are then be placed into the same group if they have the same $\mathbf{u}$, and for each value $u_k$ they have the same connectivity pattern class $P$ and incoming group $\mathbf{g}$. The exact indices $\mathbf{d}$ may be different for each neuron in the group, in which case the group can still be represented by a single connectivity function, with the various indices $\mathbf{d}$ for each neuron making the group iterable. That is, the various neurons in the group can iteratively be handled by using a loop structure to move through all neurons in the group, handling each neuron according to the exact indices that define its exact connectivity pattern. This is how a single for-loop is condensed from groups of neurons in the ENN. In general, groups of groups could be created in which each group has similar connectivity patterns to other groups, but with, for example, different $\mathbf{g}$ or only some of the same $(u_k, p^{(k)})$ pairs.

\subsubsection*{Interpreting neuron functions}
For each group, the output $y=\mathrm{sign} (b + \sum_k^m {u_k c_k} )$ is converted from this numerical function to a logical function if possible. The utility of using the condensed variables $c_k$ is not only in improving code readability, but also in making it easier to determine a simple logical function, since using condensed variables maps the input space $\mathbb{R}^m \to \mathbb{R}^{m'}$, with $m' \leq m$. There are multiple types of functions the neuron can implement, for each of which it must be checked. Several of the most pertinent function types used in this paper are described here.
\par
A simple check is used to determine if the neuron is computing a simple disjunction or conjunction (or their negated alternatives) as illustrated in Figure 1c. This is possible when it is known that the neuron's inputs $\mathbf{x}$ are discrete and bounded, such as in deeper layers where each neuron only receives connections from other hidden neurons with symbolic output. First, the condenser finds $\mathbf{x}_{\text{max}}=\mathrm{argmax}_{\mathbf{x}}{(\mathbf{w} \cdot \mathbf{x})}$. If $b+\mathbf{w} \cdot \mathbf{x}_{\text{max}}>0$, and if reducing any of the elements in $\mathbf{x}_{\text{max}}$ results in $b+\mathbf{w} \cdot \mathbf{x}'_{\text{max}}<0$, then the neuron is performing logical conjunction. Similarly, the neuron is performing logical disjunction if $b+\mathbf{w} \cdot \mathbf{x}_{\text{min}}>0$ and $b+\mathbf{w} \cdot \mathbf{x}'_{\text{min}}<0$.
\par
When there are no effective variables and all of the upstream neurons are exclusively 0s and 1s (i.e. Boolean variables), then a logical function can be found by enumerating the truth table and determining a simplified Boolean expression that produces the same truth table. To generate the truth table, the neuron's output is computed for all combinations of possible input values. To compress this truth table into a simplified Boolean formula, the Quine–McCluskey algorithm was used \cite{boolean_simplification}.
\par
The nested logic shown in Figure 1c is possible when effective variables are found or when neuron inputs can have numerous discrete values. In either case, these inputs have a larger range of discrete values they can take. When there are two such variables that have different weights, a grid can be computed with all possible combinations of values for both variables. Often it is the case that the magnitudes of these weights are quite different, leading to entire rows or columns in the grid having the same output. For example, in Fig. 1c, where the output is always FALSE when $c_2 < 3$ and always TRUE when $c_2 > 3$; the value of $c_1$ only matters when $c_2=3$. The neuron can thus be represented by a set of if-statements detailing each of these cases, some of which may be conditioned on the value of the other. By doing this, the logic of these more complicated neurons becomes more understandable, as can be seen in Figure 1c and in the distilled code for the orientation problem.
\par
\subsubsection*{Writing code}
Once the overall algorithmic structure of the ENN has been condensed in groups with logical functions, it can be translated into any desired programming language or pseudocode, or even simple sentences. The results shown in this paper demonstrate this automatic programming with Python. In any case, an initialization needs to be written that imports necessary libraries and defines a function with proper inputs. Then code is written for each group of neurons one-by-one in the order in which information passes through them. For each group, first any condensed variables are defined, computed, and stored in temporary variables. If there is only one neuron in the group, the neuron's function is written and its output assigned to some created variable. If there are multiple neurons, the outputs may be stored in a vector or matrix, and a single function will be written within the for-loop(s) that iterate over all the values in $\mathbf{d}$. Within each for-loop, any necessary condensed variables are written in, and then the logical function is written as appropriate. Once each group's functions are written in the code, a return statement is written.

\subsection*{Input size generalization}
An ENN, like other layered neural networks, can only be trained on samples of a fixed input size. However, for all of the problems we considered, varying the size of the input always gave the same size output code. Generally this is because either the code learned to ignore additional variables (such as with cellular automata) or the only difference was the range over which for-loops were iterated. In the code this manifested as the same overall code differing only in the value of certain numbers. We distilled code multiple times for each problem from data of varying input sizes and observed how these numbers changed as a function of input size. In each case these numbers followed a linear relationship, and so we were able to manually substitute these numbers with a linear function of the input size.

\subsection*{Single-rule elementary cellular automata}
Grids used for training were generated randomly, and the output label was the state of the grid's center cell at the next time step, according to the particular rule of the automaton. We have shown results for rule 30 because it famously produces chaotic patterns and for rule 110 because it is the simplest known Turing machine and because the decision tree is unable to learn it without training on all $2^n$ possible $n$-cell grids.

\subsection*{256-rule elementary cellular automata}
Grids were generated as for the single-rule case, but to this was concatenated an 8-bit string encoding the relevant rule number. For grid sizes of $n$ cells, all $2^n$ possible grids were generated for each rule. The distilled code was the same whether we trained on samples from all 256 rules on 16 specific rules. These 16 rules were 1, 2, 4, 8, 16, 64, 127, 128, 191, 223, 239, 247, 251, 253, and 254, which all have 8-bit encodings that contain either a single 1 bit or a single 0 bit.

\subsection*{Game of Life cellular automata}
For each different grid dimensions, two-dimensional grids were generated randomly for both the training and test sets, with the output label representing the state of the grid's center cell at the next time step.

\subsection*{Necessary number of training samples}
For the rule 30, rule 110, and game of life automata, we determined how many training samples were necessary for deep distilling to consistently learn the rule. This was done by randomly generating training sets with different numbers of samples and seeing how many samples were necessary to achieve perfect accuracy 10 out of 10 times. This accuracy was measured by testing on either all $2^n$ possible $n$-cell grids or on 1 million of them, whichever was less. The range is shown for 5 independent trials of this for each method.
\par
For the game of life, it was infeasible to train standard deep learning and decision trees on 5x5  grids ($2^{25} \approx \num{3.4e6}$ total possible training samples), so we instead trained and tested these on 3x3 grids with additional cells added to the perimeter so that we could test grid sizes between 9 (3x3) and 25(5x5).

\subsection*{Maximum absolute value problem}
The training data consisted of 20 values, all of which were zero except for a single number, which could be either 1 or -1, for a total of 40 training samples. It was also done with 18 and 19 values to generalize the code for input size. To test empirically how well the distilled code generalized, the code was verified to work with sets of random numbers between in the ranges [-1, 1], [-10, 10] and [-100, 100].

\subsection*{MAX-SAT problem}
The MAX-SAT training and test sets were generated as described previously \cite{enns}. Briefly, formulae were represented as matrices, rows representing clauses and columns representing variables ($x_i$ or NOT $x_i$). Training samples contained only one non-empty clause, which included either $x_1$ or NOT $x_1$ and at most one additional variable. The training set contained all possible combinations of such formulae. The ideal $x_1$ assignment was represented by 2 values for TRUE and FALSE: (1,0) for TRUE, (0,1) for FALSE, (0.99,0.01) for TRUE/FALSE, and (0.01,0.99) for FALSE/TRUE, the latter two indicating that while either TRUE or FALSE would work, one was preferred over the other. The test sets included 5000 unique formula matrices with a variable number of total possible clauses and variables. For MAX-3SAT each clause contained 3 randomly chosen variables, while for MAX-SAT each of the $n$ variables was included with probability $3/n$. Each algorithm (deep distilling, pure greedy, or 3/4-approximation) returns a probability of assigning TRUE or FALSE to $x_1$. After assignment, all satisfied clauses were removed. The variables' indices were then shifted by one and the process repeated. The results were averaged over 100 trials for each formula matrix size.

\subsection*{Shape orientation problem}
The training data consisted of 28x28 pixel images that contained a black background and a single white stripe that filled an entire row ("horizontal" label) or column ("vertical" label) of the image, for a total of 56 training samples. It was also done on 27x27 and 26x26 images to generalize the code for input size. Several data sets were used to assess the generalizability of the code to new problems. These include shorter line segments; diagonally oriented line segments; line segments made of sparse dots; zigzag lines made up of line segments at 45 degree angles; and rectangular boxes. In each case the test images' labels were assigned "horizontal" if the shape was wider than it was tall and "vertical" if it was taller than it was wide. We also generated images with low a signal-to-noise ratio (SNR) in which a line segment could be different shades of gray such that the sum of all its pixel values was equal to a preset total signal intensity level. Then speckle noise was added to the image, randomly flipping a given number of pixel values. The SNR was defined as the total intensity of the line divided by the average intensity of the noise in each row, that is $\text{SNR} = \frac{\text{total signal}}{(\text{num flipped pixels}) / \sqrt{\text{total num pixels}}}$.
\par
The Hough transform's ability to distinguish horizontal and vertical shapes was used as a point of comparison. In general the Hough transform computes the sum of pixels along lines oriented at different angles and at different angles from the origin. To distinguish between vertical and horizontal shapes, only 0\degree and 90\degree are needed. Whichever of these two angles contains the maximum value in the Hough transform is the output we use to classify the image.

\subsection*{Translating ternary neurons to binary neurons}
The neurons used in this work use the sign function as an activation function. The output of neuron $n$ is $y^{(n)}=\mathrm{sgn}(\textbf{w}^{(n)} \cdot \textbf{x}^{(n)} + b^{(n)})$. This is important because the ternary output allows ties to be explicit, such as when an input lies exactly on the hyperplane of a differentia neuron. Because $w \mathrm{sgn}(x) = w \mathbbm{1}_{x>0} - w \mathbbm{1}_{-x>0}$, it is mathematically equivalent to substitute a pair of binary neurons for the ternary neuron. One of these neurons maintains all of the original parameters (i.e. neuron bias and weights of both incoming and outgoing synapses), while the other neuron takes the negative of all these parameters.

\end{document}